\documentclass{article}
\usepackage{ijcai26}

\usepackage{times}
\usepackage{soul}
\usepackage{url}
\usepackage[hidelinks]{hyperref}
\usepackage[utf8]{inputenc}
\usepackage[small]{caption}
\usepackage{graphicx}
\usepackage{amsmath}
\usepackage{amsthm}
\usepackage{booktabs}
\usepackage{algorithm}
\usepackage{algorithmic}
\usepackage[switch]{lineno}
\usepackage{multirow}
\usepackage[table]{xcolor}
\usepackage{amssymb}

\title{Language-Agnostic Visual Embeddings for Cross-Script Handwriting Retrieval}

\author{
Fangke Chen$^{1,2}$
\and
Tianhao Dong$^3$\and
Sirry Chen$^{2,4}$\and \\
Guobin Zhang$^1$\and
Yishu Zhang$^1$\And
Yining Chen$^1$\footnote{Corresponding Author} \\
\affiliations
$^1$School of Integrated Circuits, Zhejiang University\\
$^2$Shanghai Innovation Institute\\
$^3$School of Electrical and Electronic Engineering (EEE), Nanyang Technological University\\
$^4$School of Data Science, Fudan University\\
\emails
fkchen@zju.edu.cn,
DONG0159@e.ntu.edu.sg,
siyuanchen25@m.fudan.edu.cn,
zhangguobin@zju.edu.cn,
zhangyishu@zju.edu.cn,
yining.chen@zju.edu.cn
}

\begin{document}

\maketitle

\begin{abstract}
    Handwritten word retrieval is vital for digital archives but remains challenging due to large handwriting variability and cross-lingual semantic gaps. While large vision–language models offer potential solutions, their prohibitive computational costs hinder practical edge deployment. To address this, we propose a lightweight asymmetric dual-encoder framework that learns unified, style-invariant visual embeddings. By jointly optimizing instance-level alignment and class-level semantic consistency, our approach anchors visual embeddings to language-agnostic semantic prototypes, enforcing invariance across scripts and writing styles. Experiments show that our method outperforms 28 baselines and achieves state-of-the-art accuracy on within-language retrieval benchmarks. We further conduct explicit cross-lingual retrieval, where the query language differs from the target language, to validate the effectiveness of the learned cross-lingual representations. Achieving strong performance with only a fraction of the parameters required by existing models, our framework enables accurate and resource-efficient cross-script handwriting retrieval.
\end{abstract}

\section{Introduction}

Handwriting retrieval stands as a cornerstone for mining content from massive unstructured documents, playing an indispensable role in domains such as document analysis~\cite{wolf2024self} and historical manuscript mining~\cite{perissier2024pret19}. Unlike the standardized patterns of printed text, however, handwriting processing presents significantly greater challenges due to inherent stylistic diversity and structural irregularity~\cite{zhang2019sequence}. This complexity is further exacerbated in real-world multilingual scenarios, where systems are confronted with a dual challenge. First, extreme morphological ambiguity arises from arbitrary cursive strokes and elastic distortions. Second, cross-lingual semantic gaps create substantial barriers, where synonymous concepts share no visual resemblance despite possessing identical meanings. Consequently, the fundamental objective of this task transcends simple character recognition, necessitating the construction of a semantic-invariant unified representation space capable of bridging these distinct linguistic and stylistic barriers.

\begin{figure}[t]
    \centering
    \includegraphics[width=\linewidth]{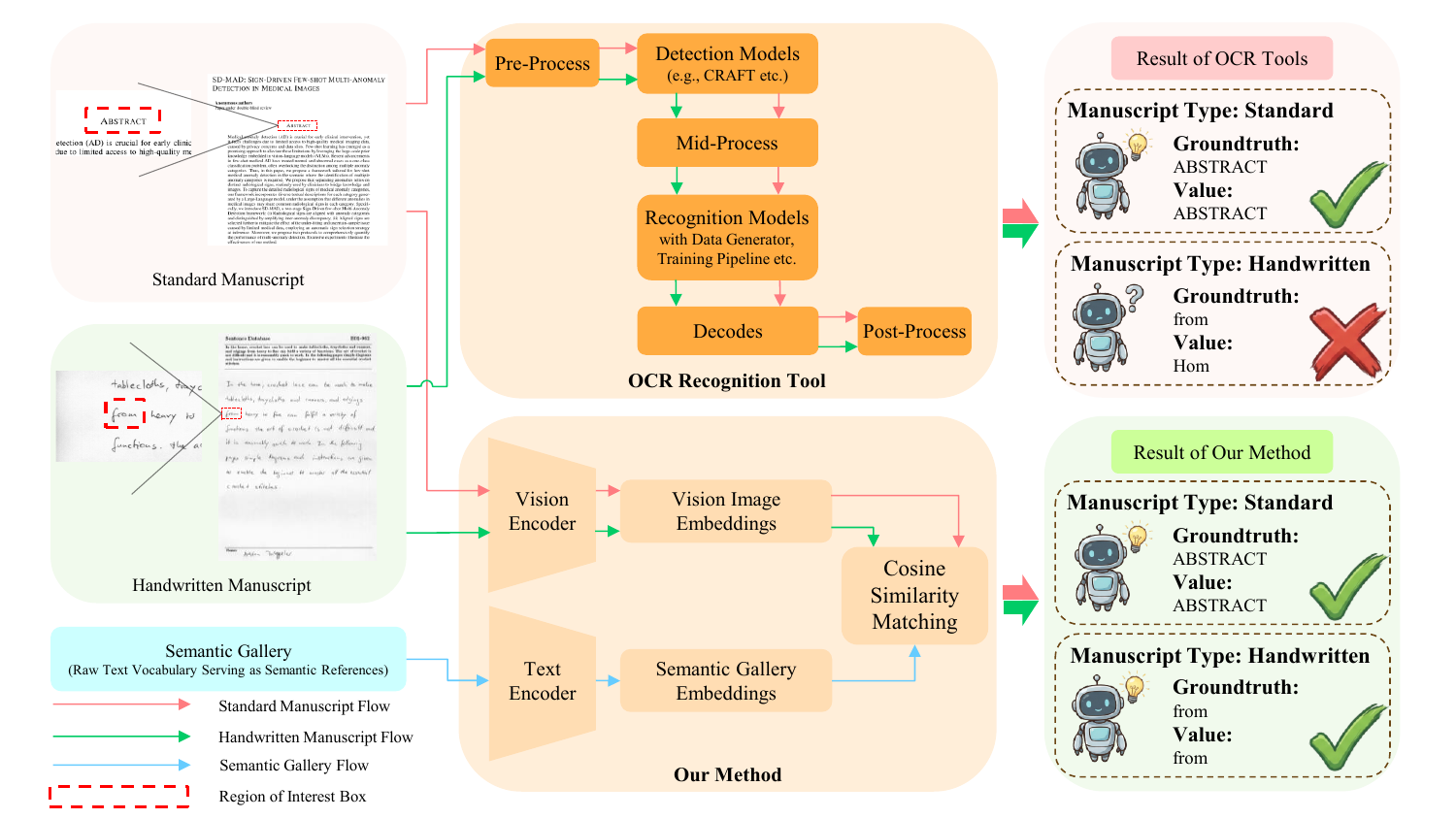} 
    \caption{Comparison of inference paradigms. Traditional OCR-based methods (top) suffer from irreversible error propagation on cursive handwriting (e.g., misrecognizing ``from'' as ``Hom''). In contrast, our method (bottom) bypasses character decoding, achieving robust retrieval by directly aligning visual and semantic embeddings in a unified space.}
    \label{fig:paradigms}
\end{figure}


Existing solutions generally fall into two paradigms. The first category comprises transcription-based methods, such as Optical Character Recognition (OCR)~\cite{Fakhet2024AnAO}, which are fundamentally constrained by irreversible error cascading: arbitrary cursive strokes frequently induce character-level misrecognition, thereby derailing the subsequent retrieval pipeline. As illustrated in Figure~\ref{fig:paradigms}, this structural fragility is particularly pronounced in cursive handwriting, where a single character misrecognition leads to a complete failure of the downstream retrieval. The second category consists of embedding-based methods, which project images into a latent space to bypass explicit transcription~\cite{mhiri2019word,kim2024deep}. However, these traditional embedding approaches typically prioritize learning visual textures or morphological similarities at the string level, lacking deep semantic understanding. Consequently, they struggle to bridge the massive orthographic gap between different writing systems to construct a unified representation space in cross-script scenarios. While recently emerging Visual Large Language Models (VLLMs) have alleviated these issues through robust semantic alignment, their prohibitive computational costs and inference latency render them unfeasible for deployment on resource-constrained edge devices. Moreover, general-purpose VLLMs often lack domain-specific sensitivity to fine-grained handwriting distortions. This creates a critical need for a solution that maintains VLLM-level semantic robustness while meeting the efficiency requirements of edge computing.

To address these challenges, we propose a lightweight asymmetric dual-encoder framework for semantic-invariant handwriting retrieval. By utilizing a partially frozen multilingual text encoder as a stable semantic anchor, our visual encoder maps high-variance handwritten images into a unified representation space, mitigating semantic drift while reducing model complexity. We jointly optimize instance-level alignment and class-level semantic consistency to encourage the separation of semantic content from language- and style-related variations. Distinct from conventional instance discrimination approaches, our method reframes handwriting retrieval as the construction of a shared, semantic-invariant topology across diverse scripts. The main contributions of this paper are summarized as follows:
\begin{itemize}
    \item We introduce an asymmetric framework utilizing frozen semantic anchors, which mitigates the risk of semantic drift in low-resource handwriting scenarios while enabling robust cross-lingual alignment.
    \item We introduce a semantic consistency alignment objective that enforces class-level semantic invariance across languages, encouraging the separation of lexical meaning from handwriting style in transcription-free cross-lingual handwriting retrieval.
    \item Experiments demonstrate that our method outperforms 28 baselines and achieves state-of-the-art (SOTA) performance on same-language handwriting retrieval tasks. Comparisons with recent VLLMs further show competitive accuracy with substantially higher efficiency.
    \item Additional cross-language retrieval experiments validate the learned cross-language representation properties of our framework, while hardware-aware simulations confirm its suitability for energy-efficient edge deployment.
\end{itemize}

\section{Related Work}
\paragraph{Handwriting Recognition and Retrieval} Handwriting analysis predominantly follows two paradigms. Transcription-based methods (OCR) decode textual content via modular pipelines~\cite{liao2017textboxes,liao2022real,shi2018aster,li2021structext}, evolving into high-performance tools~\cite{trocr,rapidocr,deepseekocr}. However, they fundamentally suffer from irreversible error propagation due to module cascading and often lack robustness against diverse handwriting styles. Conversely, embedding-based methods map images directly into a latent space to bypass explicit transcription~\cite{chineseclip,bgevisual,siglip2,gme}, significantly reducing structural complexity. Nevertheless, these approaches typically rely on language-specific priors, which severely restricts their cross-lingual generalization capabilities and adaptability to complex feature variations.

\paragraph{Cross-Lingual and Cross-Modal Representation Learning} Cross-lingual representation learning aims to align multilingual semantic concepts. Pioneering works like CLIP~\cite{clip}, ALIGN~\cite{align}, and their multilingual variants (e.g., M$^3$P~\cite{m3p}, UC$^2$~\cite{uc2}, MURAL~\cite{mural}) established the vision-language alignment paradigm via contrastive learning, extending to specialized domains~\cite{medunic}. In the scene text domain, research has focused on cross-lingual knowledge transfer or visual text synthesis~\cite{multilanguage1,multilanguage2,multilanguage3}. However, current methods primarily target natural images, lacking sensitivity to the fine-grained stroke deformations inherent in handwriting. Moreover, they generally prioritize transcription or generation over transcription-free semantic retrieval. Our framework addresses these limitations by establishing a semantic-invariant representation specifically tailored for cross-script handwriting.

\begin{figure*}[t!]
    \centering
    \includegraphics[width=\textwidth]{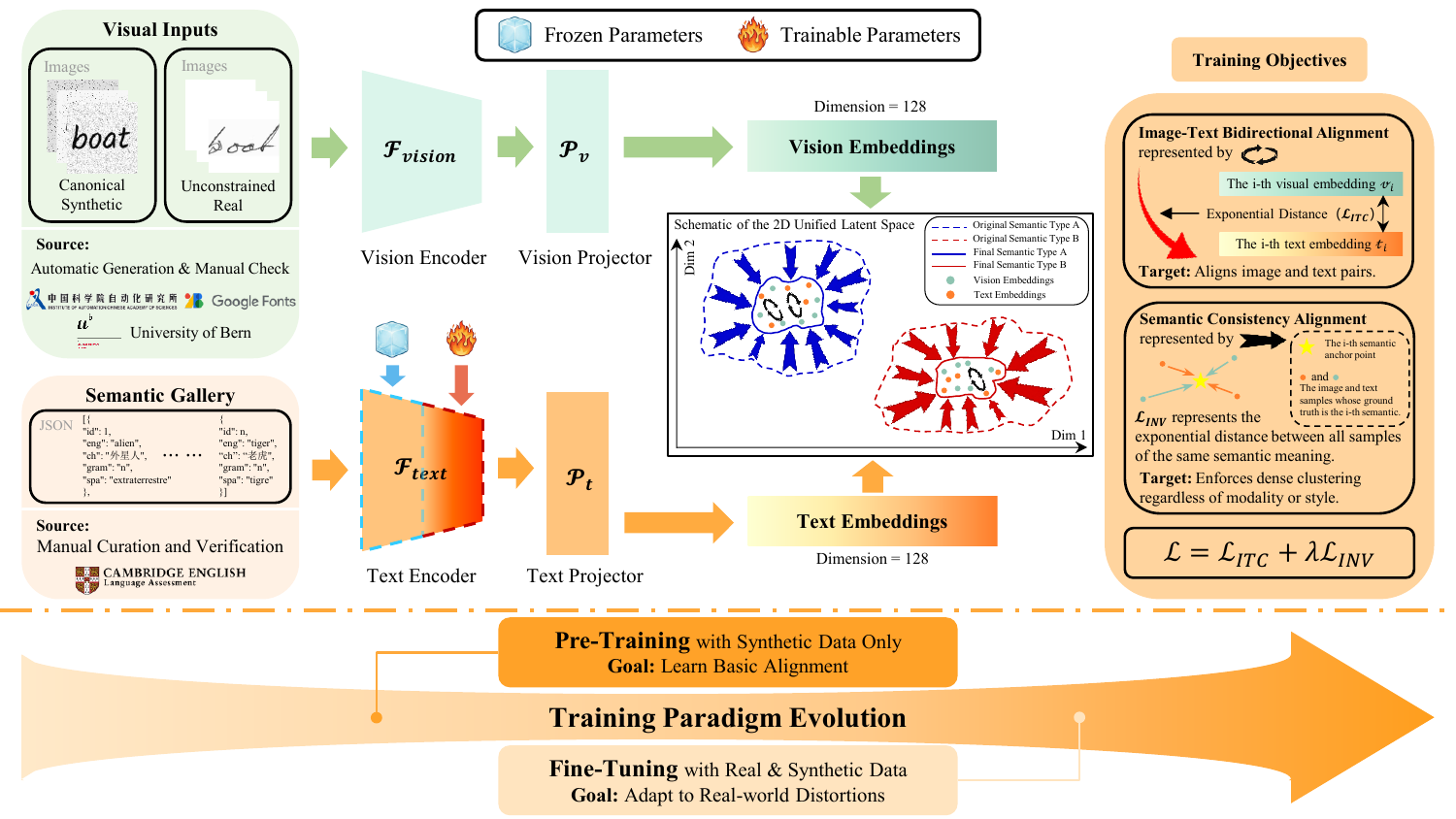} 
    \caption{The overall architecture of the proposed framework. We employ an asymmetric dual-encoder design where a frozen multilingual text encoder acts as a semantic anchor generator, and a lightweight visual encoder learns to align handwritten images with these anchors. The training process follows a progressive strategy, evolving from synthetic pre-training to real-world fine-tuning. Two complementary objectives, Instance-Level Alignment ($\mathcal{L}_{ITC}$) and Semantic Consistency Alignment ($\mathcal{L}_{INV}$), are jointly optimized to establish a unified, semantic-invariant representation space. A hyperparameter $\lambda$ (default $\lambda=0.5$) is employed to balance local discriminative power and global structural compactness.}
    \label{fig:main_framework}
\end{figure*}

\paragraph{General Visual-Language Models} With the rapid evolution of deep learning, General Visual-Language Models (GVLMs)~\cite{phi3technical,llama32vision,yao2024minicpm,chen2025janus,qwen3vltechnical,internvl35technical} have exhibited powerful capabilities in handwriting understanding driven by robust general-purpose representations. However, their prohibitive parameter counts and latency render them unfeasible for edge deployment. In contrast, our proposed method leverages a lightweight architecture to outperform these GVLMs on Out-of-Domain (OOD) benchmarks with significantly lower resource consumption, offering a practical solution for efficient handwriting retrieval.

\section{Methodology}
We first formulate the semantic-invariant handwriting retrieval problem. Subsequently, we detail the proposed lightweight asymmetric dual-encoder architecture tailored for efficient deployment, followed by the optimization objectives designed to achieve robust cross-lingual alignment and semantic consistency.

\subsection{Problem Formulation}
We aim to learn a shared, semantic-invariant representation space for cross-lingual handwriting retrieval. Formally, the multilingual dataset is defined as $N$ quadruplets $\mathcal{D} = \{(x_i, t_i, y_i, l_i)\}_{i=1}^N$. Here, $x_i \in \mathcal{X}$ and $t_i \in \mathcal{T}$ denote the handwritten image and its textual transcription, respectively. Crucially, the semantic ID is defined as a zero-based integer $y_i \in \{0, \dots, C-1\}$ acting as a language-agnostic identifier, which anchors cross-lingual synonyms to the exact same value ($C$ is the total category count). We emphasize that the semantic ID does not require sentence-level annotation or parallel corpora. In practice, it corresponds to a lightweight semantic lexicon, such as query keywords, dictionary entries, or indexed terms commonly used in document retrieval systems. This setting assumes a fixed semantic vocabulary and is standard in retrieval-based applications rather than open-ended language understanding. The tag $l_i \in \mathbb{L}$ denotes the specific language identity of the sample, which is utilized to ensure balanced language sampling.

Our goal is to learn mappings $\mathcal{F} = [f_v, f_t]$ projecting inputs into a unified space $\mathcal{V}$, yielding $L_2$-normalized embeddings $v_i = f_v(x_i)$ and $z_i = f_t(t_i)$. To facilitate retrieval, $\mathcal{V}$ must satisfy two geometric constraints based on cosine similarity $\mathcal{S}$. First, we maximize alignment for positive pairs ($y_i = y_j$) regardless of modality:
\begin{equation}
    \label{eq:alignment}
    \text{maximize} \inf_{y_i=y_j} \mathcal{S}(v_i, z_j), i,j \in \{1,2,...,N\}.
\end{equation}
Conversely, we minimize similarity for negative pairs ($y_i \neq y_j$) to ensure distinguishability:
\begin{equation}
    \label{eq:separation}
    \text{minimize} \sup_{y_i \neq y_j} \mathcal{S}(v_i, z_j), i,j \in \{1,2,...,N\}.
\end{equation}
These constraints formally define an ideal structure where intra-class variance is minimized and inter-class margin is maximized.

\subsection{Overall Framework}


As illustrated in Figure~\ref{fig:main_framework}, the framework adopts an asymmetric dual-tower architecture based on a stabilized anchor strategy. We leverage a partially frozen pre-trained text encoder $\mathcal{F}_{text}$ and a learnable projector $\mathcal{P}_t$ to construct a semantic gallery. This generates stable, language-agnostic anchors $z_i$, effectively preventing semantic collapse. What's more, for efficiency, the visual branch employs a lightweight CNN backbone $\mathcal{F}_{vision}$ with a projector $\mathcal{P}_v$. This fully trainable module maps images $x_i$ to embeddings $v_i$ by projecting variable handwritten strokes into the vicinity of the corresponding anchors. The model follows a progressive two-stage training paradigm, jointly optimized via Instance-Level Alignment ($\mathcal{L}_{ITC}$) and Semantic Consistency Alignment ($\mathcal{L}_{INV}$).

\subsection{Optimization Objectives}
To construct a geometrically compact manifold efficient for edge devices, we employ a large-batch stochastic co-occurrence strategy. By aligning the batch size to the total category magnitude, this high-density sampling ensures cross-lingual synonym pairs emerge with high probability. This yields diverse positive gradients for topological alignment without complex sampling overhead. Additionally, we prevent degenerate batches by weakly enforcing multiple instances per semantic ID. We jointly optimize instance-level and semantic consistency alignment to address cross-modal gaps and stylistic variations, respectively.

\noindent \textbf{Instance-Level Alignment.} Given a training batch $\mathcal{B} = \{(x_i, t_i, y_i, l_i)\}_{i=1}^N$, we obtain normalized embedding pairs $\{(v_i, z_i)\}_{i=1}^N$ through the visual and text branches. Note that we adopt a partial freezing strategy for the text encoder—freezing the bottom layers to preserve fundamental linguistic features while fine-tuning the upper layers to adapt to the specific handwriting semantic domain. Based on this, we employ a symmetric InfoNCE loss to establish a robust bidirectional mapping. The total contrastive loss is defined as:
\begin{equation}
    \mathcal{L}_{ITC} = \frac{1}{2} (\mathcal{L}_{V2T} + \mathcal{L}_{T2V}).
\end{equation}
We elaborate on the distinct physical interpretations of these two directions under our asymmetric architecture:
\paragraph{a) Image-to-Text Loss ($\mathcal{L}_{V2T}$): Forward Semantic Identification.}
This term drives the primary visual-to-semantic mapping. For a given visual query $v_i$, we treat the corresponding text anchor $z_i$ as the positive key and all other texts $\{z_j\}_{j \neq i}$ in the batch as negatives:
\begin{equation}
    \mathcal{L}_{V2T} = - \frac{1}{N} \sum_{i=1}^{N} \log \frac{\exp(v_i^\top z_i / \tau)}{\sum_{j=1}^{N} \exp(v_i^\top z_j / \tau)},
\end{equation}
where $\tau$ is a learnable temperature parameter. Physically, $\mathcal{L}_{V2T}$ acts as an $N$-way classification proxy, forcing the visual encoder to overcome morphological distortions and project the handwriting representation into the high-probability vicinity of its ground-truth semantic anchor, thereby ensuring accurate semantic alignment.

\paragraph{b) Text-to-Image Loss ($\mathcal{L}_{T2V}$): Inverse Discriminative Regularization.}
This term imposes an inverse constraint on the visual space. Although the text encoder serves as a stable anchor, this objective is crucial for shaping the visual distribution:
\begin{equation}
    \mathcal{L}_{T2V} = - \frac{1}{N} \sum_{i=1}^{N} \log \frac{\exp(z_i^\top v_i / \tau)}{\sum_{j=1}^{N} \exp(z_i^\top v_j / \tau)}.
\end{equation}
$\mathcal{L}_{T2V}$ serves as a structural regularizer that prevents visual collapse and enforces uniformity, significantly enhancing discriminative power against fine-grained ambiguities. With multilingual synonyms pre-aligned in the frozen textual space, cross-lingual retrieval emerges as a natural byproduct of $\mathcal{L}_{ITC}$ optimization; this implicitly coerces diverse handwritten scripts into a unified visual manifold, effectively bridging linguistic gaps without explicit supervision.


\noindent \textbf{Semantic Consistency Alignment.} The exclusive reliance on $\mathcal{L}_{ITC}$ is limited as it treats all non-diagonal samples as negatives. However, in cross-script handwriting retrieval, diverse writing systems and user styles introduce significant intra-modal variance. To enforce the model to learn semantic invariance rather than language specificity, we propose a label-guided category-level language-agnostic invariance loss ($\mathcal{L}_{INV}$).

We construct a universal feature set $\mathcal{H} = \{v\} \cup \{z\}$ containing all visual and textual embeddings in the current batch, and define an omni-modal semantic mask $M_{jk}$:
\begin{equation}
    M_{jk} = 1(y_j = y_k) \cdot 1(j \neq k).
\end{equation}
Here, $y_j = y_k$ implies that pairs sharing the same meaning, regardless of distinct languages or modalities, are treated as positives. $\mathcal{L}_{INV}$ is defined to maximize the average similarity of all positive pairs:
\begin{equation}
    \mathcal{L}_{INV} = 1 - \frac{\sum_{h_j,h_k \in \mathcal{H}} M_{jk} \cdot (h_j^\top h_k)}{\sum_{h_j,h_k \in \mathcal{H}} M_{jk} + \epsilon},
\end{equation}
where $\epsilon$ is a smoothing term. Physically, this term facilitates manifold clustering by encouraging embeddings of the same semantic class to cluster in the shared space. This process implicitly suppresses variations associated with language and writing style, leading to structural overlap of heterogeneous instances on the hypersphere. As a result, semantic content is decoupled from stylistic noise, yielding a representation topology that supports cross-lingual generalization.

\noindent \textbf{Total Objective.} Consequently, the overall training objective integrates these complementary constraints into a weighted sum:
\begin{equation}
    \mathcal{L} = \mathcal{L}_{ITC} + \lambda \cdot \mathcal{L}_{INV},
\end{equation}
where the hyperparameter $\lambda$ governs the trade-off between local discriminative power and global structural compactness. We empirically set $\lambda = 0.5$ for all experiments in this study.

\begin{table}[t!]
    \centering
    \caption{Statistics of evaluation dataset. Breakdowns of sample counts (\# Samples) and visual styles (Styles) for In-Domain and Out-of-Domain sets across languages.}
    \label{tab:ood_stats}
    \setlength{\tabcolsep}{5pt}
    \begin{tabular}{lrrrr}
        \toprule
        \textbf{Metric} & \textbf{Total} & \textbf{zh} & \textbf{en} & \textbf{es} \\
        \midrule
        \multicolumn{5}{l}{\textit{\textbf{In-Domain}}} \\
        \# Samples & 65,700 & 21,900 & 21,900 & 21,900 \\
        Styles     & 45     & 15     & 15     & 15     \\
        \midrule
        \multicolumn{5}{l}{\textit{\textbf{Out-of-Domain}}} \\
        \# Samples & 19,880 & 7,200  & 5,376  & 7,304  \\
        Styles     & $>220$ & $>100$ & $>100$ & $>20$  \\
        \bottomrule
    \end{tabular}
\end{table}

\subsection{Edge-Device Friendliness}
Unlike conventional systems where parameter size grows linearly with the number of languages, our framework compresses universal semantics into a single lightweight backbone via $\mathcal{L}_{ITC}$ and $\mathcal{L}_{INV}$. This ensures the model size remains minimal and independent of the supported language count, facilitating efficient deployment on resource-constrained edge devices.


\definecolor{bg_two_stage}{RGB}{214, 220, 228}
\definecolor{bg_end_to_end}{RGB}{221, 235, 247}
\definecolor{bg_vlm}{RGB}{186, 206, 253}
\begin{table*}[t!]
    \centering
    \caption{General retrieval performance results. We compare our framework against three categories of baselines to evaluate representation quality in within-language or mixed settings. Inference parameters and average inference latency per sample are also reported. Best scores are bolded. The symbols $\uparrow$ and $\downarrow$ indicate whether higher or lower values denote better performance.}
    \label{tab:results}
    \resizebox{\textwidth}{!}{
        \begin{tabular}{ll ccccc ccccc rr}
            \toprule
            \multirow{2}{*}{\textbf{Method}} & \multirow{2}{*}{\textbf{Source}} & \multicolumn{5}{c}{\textbf{In-Domain Set}} & \multicolumn{5}{c}{\textbf{Out-of-Domain Set}} & \multirow{2}{*}{\textbf{Paras($\downarrow$)}} & \multirow{2}{*}{\textbf{Latency($\downarrow$)}} \\
            \cmidrule(lr){3-7} \cmidrule(lr){8-12}
            & & Acc@1($\uparrow$) & Acc@3($\uparrow$) & Acc@5($\uparrow$) & MRR($\uparrow$) & NES($\uparrow$) & Acc@1($\uparrow$) & Acc@3($\uparrow$) & Acc@5($\uparrow$) & MRR($\uparrow$) & NES($\uparrow$) & (M) & (ms) \\
            \midrule
            \rowcolor{bg_two_stage} \multicolumn{14}{c}{\textit{\textbf{Two-Stage Strategies}}} \\
            EasyOCR & \cite{easyocr} & 0.8598 & 0.8872 & 0.8948 & 0.8767 & 0.8396 & 0.6044 & 0.6685 & 0.6848 & 0.6448 & 0.5982 & 30.10 & 20.33 \\
            ABINet & \cite{abinet} & 0.6485 & 0.6621 & 0.6645 & 0.6564 & 0.6365 & 0.5879 & 0.6134 & 0.6180 & 0.6023 & 0.5867 & 36.86 & 14.21 \\
            RapidOCR & \cite{rapidocr} & 0.9098 & 0.9293 & 0.9325 & 0.9210 & 0.9081 & 0.5863 & 0.6258 & 0.6344 & 0.6125 & 0.6049 & 15.00 & 96.10 \\
            PARSeq & \cite{parseq} & 0.6388 & 0.6585 & 0.6629 & 0.6503 & 0.6322 & 0.5761 & 0.6088 & 0.6165 & 0.5945 & 0.5791 & 6.02 & 10.75 \\
            TrOCR & \cite{trocr} & 0.5983 & 0.6301 & 0.6375 & 0.6163 & 0.5648 & 0.5152 & 0.5582 & 0.5703 & 0.5405 & 0.5166 & 333.92 & 29.74 \\
            Surya & \cite{surya} & 0.6331 & 0.6519 & 0.6578 & 0.6445 & 0.6270 & 0.4990 & 0.5299 & 0.5401 & 0.5177 & 0.5005 & 38.42 & 96.70 \\
            Florence-2-Large (OCR) & \cite{florence} & 0.6575 & 0.6763 & 0.6821 & 0.6773 & 0.6616 & 0.4113 & 0.4434 & 0.4509 & 0.4357 & 0.4502 & 776.47 & 159.09 \\
            DeepSeek-OCR & \cite{deepseekocr} & 0.3760 & 0.4350 & 0.4563 & 0.4183 & 0.4267 & 0.2696 & 0.3339 & 0.3610 & 0.3196 & 0.3164 & 3336.11 & 719.44 \\
            \midrule
            \rowcolor{bg_end_to_end} \multicolumn{14}{c}{\textit{\textbf{End-to-End Strategies}}} \\
            Chinese CLIP ViT Base & \multirow{3}{*}{\cite{chineseclip}} & 0.5973 & 0.7034 & 0.7359 & 0.6651 & 0.6692 & 0.3656 & 0.4559 & 0.4978 & 0.4341 & 0.4584 & 188.26 & 26.35 \\
            Chinese CLIP ViT Large &  & 0.6877 & 0.7837 & 0.8114 & 0.7481 & 0.7524 & 0.4597 & 0.5570 & 0.5977 & 0.5302 & 0.5515 & 406.23 & 20.60 \\
            Chinese CLIP ViT Huge &  & 0.6786 & 0.7587 & 0.7843 & 0.7303 & 0.7277 & 0.4580 & 0.5451 & 0.5825 & 0.5227 & 0.5407 & 957.60 & 20.63 \\
            \midrule
            AltCLIP & \cite{altclip} & 0.5726 & 0.6231 & 0.6372 & 0.6062 & 0.6139 & 0.3245 & 0.3947 & 0.4214 & 0.3775 & 0.4002 & 864.19 & 26.85 \\
            Nomic-Embed-Vision & \cite{nomicembedvision} & 0.5910 & 0.6457 & 0.6548 & 0.6257 & 0.6324 & 0.3882 & 0.4580 & 0.4766 & 0.4387 & 0.4604 & 229.68 & 4.03 \\
            E5-V & \cite{e5vvv} & 0.4332 & 0.5054 & 0.5393 & 0.4867 & 0.4861 & 0.1853 & 0.2442 & 0.2937 & 0.2479 & 0.2645 & 8355.28 & 232.67 \\
            BGE-Visualized & \cite{bgevisual} & 0.3572 & 0.4451 & 0.4830 & 0.4202 & 0.4222 & 0.1039 & 0.1698 & 0.2091 & 0.1612 & 0.1921 & 872.91 & 34.13 \\
            MM-EMBED & \cite{mmembed} & 0.5305 & 0.5987 & 0.6195 & 0.5743 & 0.5809 & 0.2977 & 0.3869 & 0.4188 & 0.3615 & 0.3807 & 8175.51 & 145.17 \\
            \midrule
            SigLIP 2 ViT-B & \multirow{4}{*}{\cite{siglip2}} & 0.4524 & 0.5431 & 0.5659 & 0.5095 & 0.5243 & 0.3113 & 0.4218 & 0.4552 & 0.3840 & 0.4112 & 375.23 & 27.67 \\
            SigLIP 2 Large &  & 0.5849 & 0.6422 & 0.6596 & 0.6232 & 0.6341 & 0.4507 & 0.5307 & 0.5563 & 0.5040 & 0.5279 & 881.53 & 26.57 \\
            SigLIP 2 So400m &  & 0.6021 & 0.6602 & 0.6763 & 0.6403 & 0.6534 & 0.4832 & 0.5489 & 0.5683 & 0.5286 & 0.5463 & 1135.67 & 27.23 \\
            SigLIP 2 Giant &  & 0.6734 & 0.7386 & 0.7574 & 0.7156 & 0.7217 & 0.5526 & 0.6286 & 0.6548 & 0.6046 & 0.6148 & 1871.39 & 23.79 \\
            \midrule
            GME-Qwen2VL-2B & \multirow{2}{*}{\cite{gme}} & 0.8874 & 0.9611 & 0.9689 & 0.9257 & 0.9337 & 0.7226 & 0.8416 & 0.8705 & 0.7907 & 0.7797 & 2208.99 & 189.43 \\
            GME-Qwen2VL-7B &  & 0.9244 & 0.9776 & 0.9814 & 0.9519 & 0.9622 & 0.7802 & 0.8729 & 0.8940 & 0.8340 & 0.8396 & 7746.38 & 197.23 \\
            \midrule
            \rowcolor{bg_vlm} \multicolumn{14}{c}{\textit{\textbf{General Visual Large Language Model Strategies}}} \\
            Phi-3.5-Vision & \cite{phi3technical} & 0.6652 & 0.6828 & 0.6888 & 0.6763 & 0.6692 & 0.4740 & 0.5091 & 0.5200 & 0.4946 & 0.5052 & 4146.62 & 854.85 \\
            Llama-3.2-11B-Vision & \cite{llama32vision} & 0.6669 & 0.6960 & 0.7030 & 0.6827 & 0.6754 & 0.4828 & 0.4997 & 0.5035 & 0.4923 & 0.4970 & 10670.22 & 717.80 \\
            MiniCPM-o 2.6 & \cite{yao2024minicpm} & 0.9629 & 0.9695 & 0.9704 & 0.9664 & 0.9645 & 0.8230 & 0.8586 & 0.8691 & 0.8436 & 0.8284 & 8099.42 & 125.13 \\
            Janus-Pro-7B & \cite{chen2025janus} & 0.6380 & 0.6577 & 0.6642 & 0.6500 & 0.6414 & 0.5153 & 0.5516 & 0.5662 & 0.5387 & 0.5390 & 7420.43 & 572.83 \\
            QWen3-VL-4B & \cite{qwen3vltechnical} & \textbf{0.9751} & 0.9791 & 0.9798 & 0.9774 & 0.9778 & 0.8444 & 0.8712 & 0.8776 & 0.8597 & 0.8502 & 4437.82 & 18.21 \\
            InternVL3.5-8B & \cite{internvl35technical} & 0.9514 & 0.9707 & 0.9740 & 0.9620 & 0.6822 & 0.8484 & 0.8848 & 0.8957 & 0.8706 & 0.6261 & 8528.32 & 188.52 \\
            \midrule
            Ours w/o FineTuning & \multirow{2}{*}{/} & 0.9738 & \textbf{1.0000} & \textbf{1.0000} & \textbf{1.0000} & \textbf{0.9806} & 0.4303 & 0.5202 & 0.5558 & 0.4953 & 0.4994 & \textbf{1.29} & \textbf{2.89} \\
            Ours & & 0.9726 & \textbf{1.0000} & \textbf{1.0000} & 0.9863 & 0.9795 & \textbf{0.8605} & \textbf{0.9537} & \textbf{0.9693} & \textbf{0.9094} & \textbf{0.8903} & \textbf{1.29} & \textbf{2.89} \\
            \bottomrule
        \end{tabular}
    }
\end{table*}

\section{Experiments and Analysis}
\subsection{Experimental Settings}
To balance edge efficiency with multilingual comprehension, we instantiate the framework using MobileNetV3-Small~\cite{mobilenetv3} and a partially frozen DistilBERT~\cite{distilbert} (bottom layers frozen) to construct a 128-dimensional shared manifold. Training follows a ``Synthetic-to-Real'' paradigm covering English (en), Chinese (zh), and Spanish (es): pre-training on 262k synthetic samples followed by fine-tuning on IAM~\cite{iamdataset} and HWDB1.0~\cite{casia}. Experiments utilize an RTX 4090 with AdamW (decaying LR: 1e-4/1e-5, 20 epochs/stage). Evaluation is strictly stratified into In-Domain and Out-of-Domain (OOD) subsets, with detailed statistics summarized in Table~\ref{tab:ood_stats}. For the Spanish OOD benchmark, due to the limited availability of reproducible modern handwriting datasets, we construct the evaluation using synthetic font styles that are strictly disjoint from the training set. This setup enforces a controlled stylistic domain shift at the font level while maintaining a standardized and accessible evaluation protocol. We note that the reported results should be interpreted as robustness to unseen font styles, rather than performance on fully in-the-wild handwriting data.

\subsection{Within-Lingual Retrieval Evaluation}
We benchmark against three cohorts: Two-Stage OCR, General Visual Embeddings, and VLLMs. Retrieval protocols are adapted per architecture—cosine similarity for embeddings and normalized edit similarity (NES) for generative models—to ensure fair comparison. We report Acc@K ($K \in \{1, 3, 5\}$), MRR, and NES, defined as follows (where $r_i$ is the rank and $\mathcal{D}$ is the Levenshtein distance):
\begin{equation}
    \label{eq:acc}
    \text{Acc}@K = \frac{1}{N} \sum_{i=1}^{N} 1(r_i \le K),
\end{equation}
\begin{equation}
    \text{MRR} = \frac{1}{N} \sum_{i=1}^{N} \frac{1}{r_i},
\end{equation}
\begin{equation}
    \label{eq:nes}
    \text{NES}(s_{pred}, s_{gt}) = 1 - \frac{\mathcal{D}(s_{pred}, s_{gt})}{\max(|s_{pred}|, |s_{gt}|)}.
\end{equation}

\begin{figure}[t!]
    \centering
    \makebox[\columnwidth][c]{%
        \includegraphics[width=1.08\columnwidth]{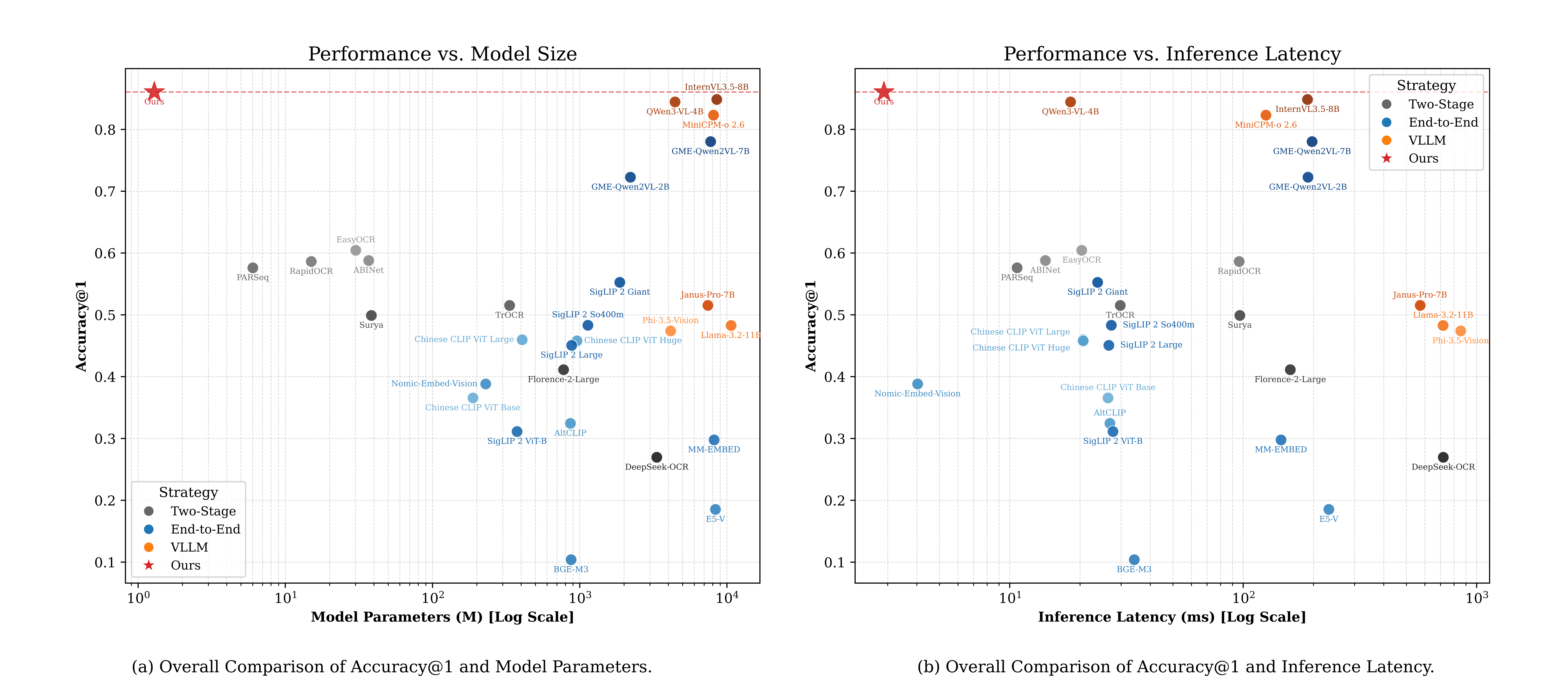}
    }
    \caption{Efficiency-Performance trade-off overview on the OOD set. (a) Accuracy@1 vs. Model Parameters. (b) Accuracy@1 vs. Inference Latency. Our framework achieves superior general retrieval performance while maintaining the minimum parameter count and lowest average latency.}
    \label{fig:comparasion}
\end{figure}

\begin{figure*}[t!]
    \centering
    \includegraphics[width=\textwidth]{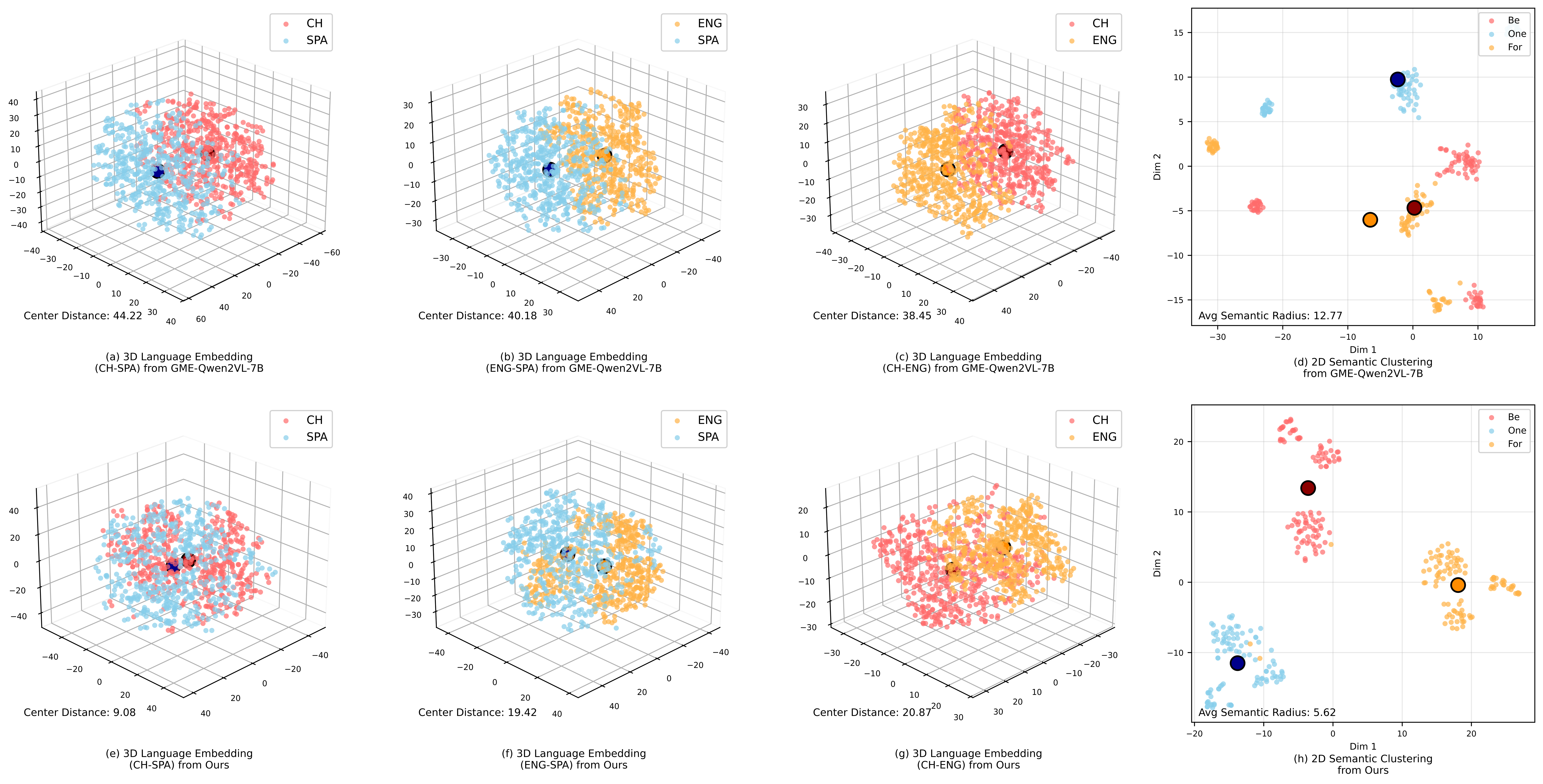} 
    \caption{Latent space visualization via t-SNE. The top row reports the embedding distribution of GME-Qwen2VL-7B, while the bottom row illustrates the results of our method. The visualization involves multilingual samples, where CH, ENG, and SPA denote Chinese, English, and Spanish data, respectively. Cluster representation centers are represented by dark-colored sample points.}
    \label{fig:visualationoflatentspace}
\end{figure*}

Table~\ref{tab:results} and Figure~\ref{fig:comparasion} present the comprehensive comparative results in within-language. Overall, our method achieves state-of-the-art performance on the challenging OOD set while maintaining highly competitive results on the In-Domain set. A universal trend where In-Domain scores consistently surpass OOD scores indicates that current models, while proficient at recognizing standardized fonts, struggle to generalize to the stylistic variance of unconstrained handwriting. Detailed analysis reveals distinct characteristics. Regarding two-stage strategies, while EasyOCR and RapidOCR demonstrate robust In-Domain performance, they falter significantly on the OOD set due to limited adaptability to handwritten stroke distortions. Furthermore, generative models like DeepSeek-OCR exhibit a substantial performance gap, likely attributed to their intrinsic deficiency in extracting fine-grained visual information from lexical images. Similarly, monolingual-centric end-to-end methods such as Chinese CLIP yield suboptimal results due to the lack of multilingual support. Notably, VLLMs have significantly advanced representation capabilities; the GME-Qwen2VL series outperforms traditional methods, while QWen3-VL-4B and InternVL3.5-8B exhibit powerful generalization comparable to our framework. Nevertheless, their prohibitive parameter counts and inference latency create substantial barriers for edge deployment. Consequently, as illustrated in Figure~\ref{fig:comparasion}, our method achieves the optimal balance between representation quality and computational efficiency.

\subsection{Cross-Lingual Retrieval Evaluation}
\begin{table}[t!]
    \centering
    \caption{Explicit cross-lingual retrieval accuracy@1(\%) on the OOD handwriting set. We report the Image-to-Text performance across different language pairs to verify the semantic invariance of the learned representations.}
    \label{tab:cross_lingual_ood}
    \resizebox{1.0\columnwidth}{!}{
        \begin{tabular}{lccccccc}
            \toprule
            \multirow{2}{*}{\textbf{Method}} & \multicolumn{6}{c}{\textbf{Image $\to$ Text}} & \multirow{2}{*}{\textbf{Avg.}} \\
            \cmidrule(lr){2-7}
             & en$\to$zh & zh$\to$en & zh$\to$es & es$\to$zh & es$\to$en & en$\to$es &  \\
            \midrule
            Random & 0.34 & 0.26 & 0.29 & 0.30 & 0.40 & 0.30 & 0.32 \\
            Chinese CLIP ViT Huge & 13.79 & 9.33 & 2.06 & 12.61 & 23.33 & 2.66 & 10.63 \\
            SigLIP 2 So400m & 27.93 & 1.68 & 1.32 & 27.97 & 41.25 & 24.67 & 20.80 \\
            SigLIP 2 Giant & 36.89 & 6.71 & 8.26 & 29.45 & 52.59 & 31.58 & 27.55 \\
            GME-Qwen2VL-2B & 30.53 & 48.79 & 30.10 & 33.20 & 42.40 & 17.58 & 33.77 \\
            GME-Qwen2VL-7B & 42.05 & 57.36 & 44.63 & 32.26 & 50.42 & 30.62 & 42.89 \\
            \midrule
            \textbf{Ours} & \textbf{73.55} & \textbf{84.96} & \textbf{83.88} & \textbf{90.36} & \textbf{90.98} & \textbf{73.66} & \textbf{82.80} \\
            \bottomrule
        \end{tabular}
    }
\end{table}
To further substantiate the cross-lingual representation capability of our framework, we conducted an explicit cross-lingual retrieval evaluation on the OOD set. For baselines, we selected end-to-end methods that demonstrated superior performance in the within-language evaluation. Table~\ref{tab:cross_lingual_ood} reports the accuracy@1 results. We observe that the selected baselines generally struggle with cross-lingual retrieval tasks. Even the VLLM-based GME-QwenVL series fails to exhibit robust cross-lingual representation capabilities, while Chinese CLIP ViT Huge, which relies on language-specific visual patterns, shows almost negligible cross-lingual capability. In contrast, our framework delivers significantly superior results, further validating the effectiveness of its cross-lingual representation. These results indicate that our framework successfully decouples semantics from language-specific styles, establishing a truly language-agnostic embedding space.

\subsection{Characterization Analysis}
To investigate the geometric mechanisms underlying the superior performance, we evaluate the latent space topology. To mitigate scale variations arising from differing embedding dimensions, we employ the normalized metric $R/D = R_{intra} / D_{inter}$, defined as the ratio of Mean Intra-Class Radius ($R_{intra}$) to Mean Inter-Class Distance ($D_{inter}$). A lower ratio indicates a more compact and separable manifold. As shown in Table~\ref{tab:characterization}, our method achieves the lowest R/D Ratio, outperforming parameter-heavy VLM-based GME-Qwen2VL series and suggesting superior geometric structuring over brute-force fitting. This is corroborated by visualizations in Figure~\ref{fig:visualationoflatentspace}: our method drastically shortens inter-lingual center distances (e.g., Chinese-Spanish: $44.22 \to 9.08$) via $\mathcal{L}_{ITC}$, and compresses the average semantic radius ($12.77 \to 5.62$) via $\mathcal{L}_{INV}$. These results confirm that our framework effectively bridges linguistic distributions while suppressing stylistic variance.

\begin{table}[t!]
    \centering
    \caption{Characterization analysis of the embedding space on the OOD Set. We report the Intra-Class Radius ($R_{intra} \downarrow$), Inter-Class Distance ($D_{inter} \uparrow$), and the normalized R/D Ratio ($\downarrow$). The best results for Acc@1 and R/D Ratio are highlighted in bold.}
    \label{tab:characterization}
    \setlength{\tabcolsep}{3pt}
    \resizebox{\columnwidth}{!}{
        \begin{tabular}{lcccc}
            \toprule
            \textbf{Method} & \textbf{$R_{intra} \downarrow$} & \textbf{$D_{inter} \uparrow$} & \textbf{Acc@1 $\uparrow$} & \textbf{R/D Ratio $\downarrow$} \\
            \midrule
            Chinese CLIP ViT Base & 0.6543 & 0.3996 & 0.3656 & 1.6374 \\
            Chinese CLIP ViT Large & 0.7041 & 0.5003 & 0.4597 & 1.4074 \\
            Chinese CLIP ViT Huge & 0.7591 & 0.5827 & 0.4580 & 1.3027 \\
            AltCLIP & 0.7656 & 0.4844 & 0.3245 & 1.5805 \\
            Nomic-Embed-Vision-v1.5 & 0.4767 & 0.2291 & 0.3882 & 2.0808 \\
            E5-V & 0.6559 & 0.4324 & 0.1853 & 1.5169 \\
            BGE-M3 & 0.6111 & 0.3018 & 0.1039 & 2.0249 \\
            MM-EMBED & 0.7736 & 0.5246 & 0.2977 & 1.4746 \\
            SigLIP 2 ViT-B & 0.6039 & 0.3981 & 0.3113 & 1.5170 \\
            SigLIP 2 Large & 0.6796 & 0.4592 & 0.4507 & 1.4799 \\
            SigLIP 2 So400m & 0.7021 & 0.4788 & 0.4832 & 1.4664 \\
            SigLIP 2 Giant & 0.6789 & 0.4971 & 0.5526 & 1.3657 \\
            GME-Qwen2VL-2B & 0.8220 & 0.7318 & 0.7226 & 1.1233 \\
            GME-Qwen2VL-7B & 0.8448 & 0.7465 & 0.7802 & 1.1317 \\
            \midrule
            \textbf{Ours} & 0.4062 & 0.4559 & \textbf{0.8605} & \textbf{0.8910} \\
            \bottomrule
        \end{tabular}
    }
\end{table}

\begin{table}[t!]
    \centering
    \caption{Ablation study on the OOD Set. We evaluate the impact of different objectives and the Fine-Tuning (FT) strategy. $\checkmark$ in the FT column denotes the deployment of the Real-world Fine-tuning stage. Best scores are highlighted in bold.}
    \label{tab:ablation}
    \footnotesize 
    \setlength{\tabcolsep}{0pt} 
    \begin{tabular*}{\columnwidth}{@{\extracolsep{\fill}} cccc ccc}
        \toprule
        \multicolumn{3}{c}{\textbf{Learning Objectives}} & \textbf{Strategy} & \multicolumn{3}{c}{\textbf{OOD Metrics}} \\
        \cmidrule(r){1-3} \cmidrule(lr){4-4} \cmidrule(l){5-7}
        $\mathcal{L}_{V2T}$ & $\mathcal{L}_{T2V}$ & $\mathcal{L}_{INV}$ & FT & Acc@1$\uparrow$ & MRR$\uparrow$ & NES$\uparrow$ \\
        \midrule
        \checkmark & & & & 0.4111 & 0.4682 & 0.4855 \\
        \checkmark & & & \checkmark & 0.8039 & 0.8699 & 0.8421 \\
        \checkmark & & \checkmark & & 0.4063 & 0.4601 & 0.4805 \\
        \checkmark & & \checkmark & \checkmark & 0.8060 & 0.8704 & 0.8442 \\
        & \checkmark & & & 0.2595 & 0.3220 & 0.3509 \\
        & \checkmark & & \checkmark & 0.4327 & 0.5481 & 0.5306 \\
        & \checkmark & \checkmark & & 0.3810 & 0.4371 & 0.4689 \\
        & \checkmark & \checkmark & \checkmark & 0.7289 & 0.8146 & 0.7793 \\
        & & \checkmark & & 0.0029 & 0.0094 & 0.1186 \\
        & & \checkmark & \checkmark & 0.0043 & 0.0108 & 0.1197 \\
        \checkmark & \checkmark & & & 0.4037 & 0.4605 & 0.4780 \\
        \checkmark & \checkmark & & \checkmark & 0.8348 & 0.8922 & 0.8693 \\
        \checkmark & \checkmark & \checkmark & & 0.4303 & 0.4953 & 0.4994 \\
        \checkmark & \checkmark & \checkmark & \textbf{\checkmark} & \textbf{0.8605} & \textbf{0.9094} & \textbf{0.8903} \\
        \bottomrule
    \end{tabular*}
\end{table}

\subsection{Ablation Study}
To verify the efficacy of individual components and the proposed training strategy, we conducted ablation experiments on the OOD benchmark, with results detailed in Table~\ref{tab:ablation}. First, the unidirectional Image-to-Text loss ($\mathcal{L}_{V2T}$) serves as the primary driver for establishing retrieval capabilities, yielding an Acc@1 of 0.8039 after fine-tuning. In contrast, the Text-to-Image loss ($\mathcal{L}_{T2V}$) shows weaker standalone performance, aligning with its role as a regularizer; however, combining both into a bidirectional $\mathcal{L}_{ITC}$ significantly boosts performance to 0.8348, demonstrating the importance of bidirectional alignment in shaping a uniform latent space. Furthermore, while $\mathcal{L}_{INV}$ alone cannot establish cross-modal mappings, its integration with $\mathcal{L}_{ITC}$ effectively aggregates synonymous samples, enabling the full model to achieve a peak Acc@1 of 0.8605. This validates the complementary effect where $\mathcal{L}_{ITC}$ aligns modalities while $\mathcal{L}_{INV}$ refines the manifold structure.

Regarding the training strategy, models relying solely on synthetic pre-training exhibit limited OOD performance. The introduction of real-world fine-tuning triggers a substantial performance leap across all configurations. This significant gain strongly validates the efficacy of our ``Synthetic Pre-training to Real-world Fine-tuning'' paradigm, suggesting that synthetic data establishes the topological skeleton of semantic anchors, while real data populates the detailed textures required to handle stylistic diversity, thereby ensuring robust generalization to unseen handwriting styles.

\subsection{Hardware-Aware Simulations}
\begin{figure}[t!]
    \centering
    \includegraphics[width=\columnwidth]{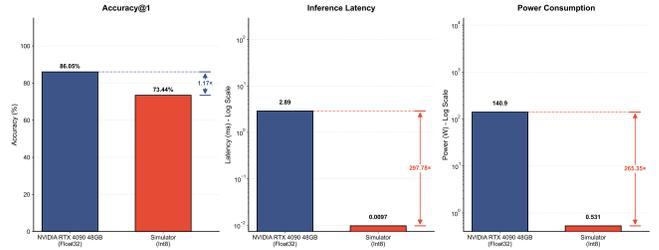} 
    \caption{Performance comparison between commercial GPU and our simulator. We report the trade-offs in accuracy, latency, and power. The \textcolor{blue}{blue arrow} indicates accuracy drop due to quantization, while \textcolor{red}{red arrows} indicate significant improvements in efficiency.}
    \label{fig:hardware_results}
\end{figure}

To verify hardware feasibility, we utilized a NeuRRAM-based compute-in-memory simulator to emulate edge inference, focusing on relative efficiency trends rather than cycle-accurate estimation. Benchmarking the int8-quantized model against a float32 baseline on an NVIDIA RTX 4090 reveals a substantial trade-off: as shown in Figure~\ref{fig:hardware_results}, despite a moderate accuracy drop, quantization unlocks immense gains, achieving $297.78\times$ latency reduction and $265.35\times$ power savings. These results confirm the framework's strong potential for resource-constrained deployment.

\section{Conclusion}
In this study, we present an edge-device-friendly framework for cross-lingual semantic invariant handwriting representation. Extensive experiments demonstrate that our approach achieves an optimal trade-off between retrieval performance and deployment costs. Characterization analysis reveals that geometric constraints significantly enhance the latent space topology. Furthermore, ablation studies verify the efficacy of our hybrid learning objectives and the two-stage training paradigm. Finally, hardware-aware simulations substantiate the feasibility and potential of our method for resource-constrained edge deployment.

\newpage

\appendix
\bibliographystyle{named}
\bibliography{ijcai26}

@inproceedings{zhang2019sequence,
  title={Sequence-to-sequence domain adaptation network for robust text image recognition},
  author={Zhang, Yaping and Nie, Shuai and Liu, Wenju and Xu, Xing and Zhang, Dongxiang and Shen, Heng Tao},
  booktitle={Proceedings of the IEEE/CVF conference on computer vision and pattern recognition},
  pages={2740--2749},
  year={2019}
}

@article{wolf2024self,
  title={Self-training for handwritten word recognition and retrieval},
  author={Wolf, Fabian and Fink, Gernot A},
  journal={International Journal on Document Analysis and Recognition (IJDAR)},
  volume={27},
  number={3},
  pages={225--244},
  year={2024},
  publisher={Springer}
}

@inproceedings{perissier2024pret19,
  title={PRET19: Automatic Recognition and Indexing of Handwritten Loan Registers from 19th Century Parisian Universities},
  author={P{\'e}rissier, L{\'e}a and Rebolledo-Dhuin, Viera and Petiot, Marie-Th{\'e}r{\`e}se and Schneider, Yoann and Kermorvant, Christopher},
  booktitle={International Conference on Theory and Practice of Digital Libraries},
  pages={360--378},
  year={2024},
  organization={Springer}
}

@article{Fakhet2024AnAO,
  title={An Arabic OCR Approach Using Levenshtein Distance and CNNs},
  author={Walid Fakhet and Salim El Khediri and Salah Zidi},
  journal={Ing{\'e}nierie des syst{\`e}mes d information},
  year={2024},
  url={https://api.semanticscholar.org/CorpusID:268147868}
}

@article{mhiri2019word,
  title={Word spotting and recognition via a joint deep embedding of image and text},
  author={Mhiri, Mohamed and Desrosiers, Christian and Cheriet, Mohamed},
  journal={Pattern Recognition},
  volume={88},
  pages={312--320},
  year={2019},
  publisher={Elsevier}
}

@article{kim2024deep,
  title={A deep learning approach for the comparison of handwritten documents using latent feature vectors},
  author={Kim, Juhyeon and Park, Soyoung and Carriquiry, Alicia},
  journal={Statistical Analysis and Data Mining: The ASA Data Science Journal},
  volume={17},
  number={1},
  pages={e11660},
  year={2024},
  publisher={Wiley Online Library}
}

@inproceedings{liao2017textboxes,
  title={Textboxes: A fast text detector with a single deep neural network},
  author={Liao, Minghui and Shi, Baoguang and Bai, Xiang and Wang, Xinggang and Liu, Wenyu},
  booktitle={Proceedings of the AAAI conference on artificial intelligence},
  volume={31},
  number={1},
  year={2017}
}

@article{liao2022real,
  title={Real-time scene text detection with differentiable binarization and adaptive scale fusion},
  author={Liao, Minghui and Zou, Zhisheng and Wan, Zhaoyi and Yao, Cong and Bai, Xiang},
  journal={IEEE transactions on pattern analysis and machine intelligence},
  volume={45},
  number={1},
  pages={919--931},
  year={2022},
  publisher={IEEE}
}

@article{shi2018aster,
  title={Aster: An attentional scene text recognizer with flexible rectification},
  author={Shi, Baoguang and Yang, Mingkun and Wang, Xinggang and Lyu, Pengyuan and Yao, Cong and Bai, Xiang},
  journal={IEEE transactions on pattern analysis and machine intelligence},
  volume={41},
  number={9},
  pages={2035--2048},
  year={2018},
  publisher={IEEE}
}

@inproceedings{li2021structext,
  title={Structext: Structured text understanding with multi-modal transformers},
  author={Li, Yulin and Qian, Yuxi and Yu, Yuechen and Qin, Xiameng and Zhang, Chengquan and Liu, Yan and Yao, Kun and Han, Junyu and Liu, Jingtuo and Ding, Errui},
  booktitle={Proceedings of the 29th ACM international conference on multimedia},
  pages={1912--1920},
  year={2021}
}

@inproceedings{trocr,
  title={Trocr: Transformer-based optical character recognition with pre-trained models},
  author={Li, Minghao and Lv, Tengchao and Chen, Jingye and Cui, Lei and Lu, Yijuan and Florencio, Dinei and Zhang, Cha and Li, Zhoujun and Wei, Furu},
  booktitle={Proceedings of the AAAI conference on artificial intelligence},
  volume={37},
  number={11},
  pages={13094--13102},
  year={2023}
}

@misc{rapidocr,
    title={{Rapid OCR}: OCR Toolbox},
    author={RapidAI Team},
    howpublished = {\url{https://github.com/RapidAI/RapidOCR}},
    year={2021}
}

@article{deepseekocr,
  title={Deepseek-ocr: Contexts optical compression},
  author={Wei, Haoran and Sun, Yaofeng and Li, Yukun},
  journal={arXiv preprint arXiv:2510.18234},
  year={2025}
}

@article{chineseclip,
  title={Chinese clip: Contrastive vision-language pretraining in chinese},
  author={Yang, An and Pan, Junshu and Lin, Junyang and Men, Rui and Zhang, Yichang and Zhou, Jingren and Zhou, Chang},
  journal={arXiv preprint arXiv:2211.01335},
  year={2022}
}

@article{siglip2,
  title={Siglip 2: Multilingual vision-language encoders with improved semantic understanding, localization, and dense features},
  author={Tschannen, Michael and Gritsenko, Alexey and Wang, Xiao and Naeem, Muhammad Ferjad and others},
  journal={arXiv preprint arXiv:2502.14786},
  year={2025}
}

@inproceedings{gme,
  title={Bridging Modalities: Improving Universal Multimodal Retrieval by Multimodal Large Language Models},
  author={Zhang, Xin and Zhang, Yanzhao and Xie, Wen and Li, Mingxin and Dai, Ziqi and Long, Dingkun and Xie, Pengjun and Zhang, Meishan and Li, Wenjie and Zhang, Min},
  booktitle={Proceedings of the Computer Vision and Pattern Recognition Conference},
  pages={9274--9285},
  year={2025}
}

@inproceedings{bgevisual,
    title = "{VISTA}: Visualized Text Embedding For Universal Multi-Modal Retrieval",
    author = "Zhou, Junjie  and
      Liu, Zheng  and
      Xiao, Shitao  and
      Zhao, Bo  and
      Xiong, Yongping",
    editor = "Ku, Lun-Wei  and
      Martins, Andre  and
      Srikumar, Vivek",
    booktitle = "Proceedings of the 62nd Annual Meeting of the Association for Computational Linguistics (Volume 1: Long Papers)",
    month = aug,
    year = "2024",
    address = "Bangkok, Thailand",
    publisher = "Association for Computational Linguistics",
    url = "https://aclanthology.org/2024.acl-long.175/",
    doi = "10.18653/v1/2024.acl-long.175",
    pages = "3185--3200"
}

@inproceedings{clip,
  title={Learning transferable visual models from natural language supervision},
  author={Radford, Alec and Kim, Jong Wook and Hallacy, Chris and Ramesh, Aditya and Goh, Gabriel and Agarwal, Sandhini and Sastry, Girish and Askell, Amanda and Mishkin, Pamela and Clark, Jack and others},
  booktitle={International conference on machine learning},
  pages={8748--8763},
  year={2021},
  organization={PmLR}
}

@inproceedings{align,
  title={Scaling up visual and vision-language representation learning with noisy text supervision},
  author={Jia, Chao and Yang, Yinfei and Xia, Ye and Chen, Yi-Ting and Parekh, Zarana and Pham, Hieu and Le, Quoc and Sung, Yun-Hsuan and Li, Zhen and Duerig, Tom},
  booktitle={International conference on machine learning},
  pages={4904--4916},
  year={2021},
  organization={PMLR}
}

@inproceedings{m3p,
  title={M3p: Learning universal representations via multitask multilingual multimodal pre-training},
  author={Ni, Minheng and Huang, Haoyang and Su, Lin and Cui, Edward and Bharti, Taroon and Wang, Lijuan and Zhang, Dongdong and Duan, Nan},
  booktitle={Proceedings of the IEEE/CVF conference on computer vision and pattern recognition},
  pages={3977--3986},
  year={2021}
}

@inproceedings{uc2,
  title={Uc2: Universal cross-lingual cross-modal vision-and-language pre-training},
  author={Zhou, Mingyang and Zhou, Luowei and Wang, Shuohang and Cheng, Yu and Li, Linjie and Yu, Zhou and Liu, Jingjing},
  booktitle={Proceedings of the IEEE/CVF Conference on Computer Vision and Pattern Recognition},
  pages={4155--4165},
  year={2021}
}

@inproceedings{mural,
    title = "{MURAL}: Multimodal, Multitask Representations Across Languages",
    author = "Jain, Aashi  and
      Guo, Mandy  and
      Srinivasan, Krishna  and
     others",
    booktitle = "Findings of the Association for Computational Linguistics: EMNLP 2021",
    month = nov,
    year = "2021",
    address = "Punta Cana, Dominican Republic",
    publisher = "Association for Computational Linguistics",
    pages = "3449--3463"
}

@article{medunic,
  title={Med-unic: Unifying cross-lingual medical vision-language pre-training by diminishing bias},
  author={Wan, Zhongwei and Liu, Che and Zhang, Mi and Fu, Jie and Wang, Benyou and Cheng, Sibo and Ma, Lei and Quilodr{\'a}n-Casas, C{\'e}sar and Arcucci, Rossella},
  journal={Advances in Neural Information Processing Systems},
  volume={36},
  pages={56186--56197},
  year={2023}
}

@inproceedings{multilanguage1,
  title={Cross-Lingual Learning in Multilingual Scene Text Recognition},
  author={Baek, Jeonghun and Matsui, Yusuke and Aizawa, Kiyoharu},
  booktitle={ICASSP 2024-2024 IEEE International Conference on Acoustics, Speech and Signal Processing (ICASSP)},
  pages={6770--6774},
  year={2024},
  organization={IEEE}
}

@inproceedings{multilanguage2,
  title={Cross-Lingual Learning for Low-Resource Khmer Scene Text Detection and Recognition},
  author={Nom, Vannkinh and Keo, Saly and Bakkali, Souhail and Luqman, Muhammad Muzzamil and Coustaty, Micka{\"e}l and Ogier, Jean-Marc},
  booktitle={ICDAR 2025 Workshop on Documents Analysis of Low-resource Languages},
  year={2025}
}

@inproceedings{multilanguage3,
  title={Cross-lingual Visual Text Stylization and Synthesis Incorporating Text Rendering and Diffusion Model},
  author={Shen, Minmin and Chen, Caren},
  booktitle={Proceedings of the IEEE/CVF International Conference on Computer Vision},
  pages={6049--6057},
  year={2025}
}

@misc{phi3technical,
      title={Phi-3 Technical Report: A Highly Capable Language Model Locally on Your Phone}, 
      author={Marah Abdin and Jyoti Aneja and Hany Awadalla and others},
      year={2024},
      eprint={2404.14219},
      archivePrefix={arXiv},
      primaryClass={cs.CL},
      url={https://arxiv.org/abs/2404.14219}, 
}

@misc{qwen3vltechnical,
      title={Qwen3-VL Technical Report}, 
      author={Shuai Bai and Yuxuan Cai and Ruizhe Chen and others},
      year={2025},
      eprint={2511.21631},
      archivePrefix={arXiv},
      primaryClass={cs.CV},
      url={https://arxiv.org/abs/2511.21631}, 
}

@misc{internvl35technical,
      title={InternVL3.5: Advancing Open-Source Multimodal Models in Versatility, Reasoning, and Efficiency}, 
      author={Weiyun Wang and Zhangwei Gao and Lixin Gu and others},
      year={2025},
      eprint={2508.18265},
      archivePrefix={arXiv},
      primaryClass={cs.CV},
      url={https://arxiv.org/abs/2508.18265}, 
}

@article{chen2025janus,
  title={Janus-pro: Unified multimodal understanding and generation with data and model scaling},
  author={Chen, Xiaokang and Wu, Zhiyu and Liu, Xingchao and Pan, Zizheng and Liu, Wen and Xie, Zhenda and Yu, Xingkai and Ruan, Chong},
  journal={arXiv preprint arXiv:2501.17811},
  year={2025}
}

@article{yao2024minicpm,
  title={Minicpm-v: A gpt-4v level mllm on your phone},
  author={Yao, Yuan and Yu, Tianyu and Zhang, Ao and Wang, Chongyi and Cui, Junbo and Zhu, Hongji and Cai, Tianchi and Li, Haoyu and Zhao, Weilin and He, Zhihui and others},
  journal={arXiv preprint arXiv:2408.01800},
  year={2024}
}

@misc{llama32vision,
  title        = {Llama 3.2 11B Vision Instruct Model – Azure AI Model Catalog},
  author       = {Meta AI},
  year         = {2024},
  url          = {https://ai.azure.com/catalog/models/Llama-3.2-11B-Vision-Instruct}
}

@inproceedings{mobilenetv3,
  title={Searching for mobilenetv3},
  author={Howard, Andrew and Sandler, Mark and Chu, Grace and Chen, Liang-Chieh and Chen, Bo and Tan, Mingxing and Wang, Weijun and Zhu, Yukun and Pang, Ruoming and Vasudevan, Vijay and others},
  booktitle={Proceedings of the IEEE/CVF international conference on computer vision},
  pages={1314--1324},
  year={2019}
}

@article{distilbert,
  title={DistilBERT, a distilled version of BERT: smaller, faster, cheaper and lighter},
  author={Sanh, Victor and Debut, Lysandre and Chaumond, Julien and Wolf, Thomas},
  journal={arXiv preprint arXiv:1910.01108},
  year={2019}
}

@article{iamdataset,
  title={The IAM-database: an English sentence database for offline handwriting recognition},
  author={Marti, U-V and Bunke, Horst},
  journal={International journal on document analysis and recognition},
  volume={5},
  number={1},
  pages={39--46},
  year={2002},
  publisher={Springer}
}

@inproceedings{casia,
  title={CASIA online and offline Chinese handwriting databases},
  author={Liu, Cheng-Lin and Yin, Fei and Wang, Da-Han and Wang, Qiu-Feng},
  booktitle={2011 international conference on document analysis and recognition},
  pages={37--41},
  year={2011},
  organization={IEEE}
}

@misc{mmembed,
      title={MM-Embed: Universal Multimodal Retrieval with Multimodal LLMs}, 
      author={Sheng-Chieh Lin and Chankyu Lee and Mohammad Shoeybi and Jimmy Lin and Bryan Catanzaro and Wei Ping},
      year={2024},
      eprint={2411.02571},
      archivePrefix={arXiv},
      primaryClass={cs.CL},
      url={https://arxiv.org/abs/2411.02571},
}

@article{e5vvv,
  title={E5-v: Universal embeddings with multimodal large language models},
  author={Jiang, Ting and Song, Minghui and Zhang, Zihan and Huang, Haizhen and Deng, Weiwei and Sun, Feng and Zhang, Qi and Wang, Deqing and Zhuang, Fuzhen},
  journal={arXiv preprint arXiv:2407.12580},
  year={2024}
}

@misc{nomicembedvision,
      title={Nomic Embed Vision: Expanding the Latent Space}, 
      author={Zach Nussbaum and Brandon Duderstadt and Andriy Mulyar},
      year={2024},
      eprint={2406.18587},
      archivePrefix={arXiv},
      primaryClass={cs.CV},
      url={https://arxiv.org/abs/2406.18587}, 
}

@inproceedings{altclip,
    title = "{A}lt{CLIP}: Altering the Language Encoder in {CLIP} for Extended Language Capabilities",
    author = "Chen, Zhongzhi  and
      Liu, Guang  and
      Zhang, Bo-Wen  and
      Yang, Qinghong  and
      Wu, Ledell",
    editor = "Rogers, Anna  and
      Boyd-Graber, Jordan  and
      Okazaki, Naoaki",
    booktitle = "Findings of the Association for Computational Linguistics: ACL 2023",
    month = jul,
    year = "2023",
    address = "Toronto, Canada",
    publisher = "Association for Computational Linguistics",
    url = "https://aclanthology.org/2023.findings-acl.552/",
    doi = "10.18653/v1/2023.findings-acl.552",
    pages = "8666--8682"
}

@article{abinet,
  title={Read Like Humans: Autonomous, Bidirectional and Iterative Language Modeling for Scene Text Recognition},
  author={Fang, Shancheng and Xie, Hongtao and Wang, Yuxin and Mao, Zhendong and Zhang, Yongdong},
    booktitle={Proceedings of the IEEE/CVF Conference on Computer Vision and Pattern Recognition},
  year={2021}
}

@misc{easyocr,
  author = {{Jaided AI}},
  title = {{EasyOCR}},
  howpublished = {\url{https://github.com/JaidedAI/EasyOCR}},
}

@inproceedings{parseq,
  title={Scene Text Recognition with Permuted Autoregressive Sequence Models},
  author={Bautista, Darwin and Atienza, Rowel},
  booktitle={European Conference on Computer Vision},
  pages={178--196},
  month={10},
  year={2022},
  publisher={Springer Nature Switzerland},
  address={Cham},
  doi={10.1007/978-3-031-19815-1_11},
  url={https://doi.org/10.1007/978-3-031-19815-1_11}
}

@misc{surya,
  author       = {Vikas Paruchuri and Datalab Team},
  title        = {Surya: A lightweight document OCR and analysis toolkit},
  year         = {2025},
  howpublished = {\url{https://github.com/VikParuchuri/surya}},
  note         = {GitHub repository},
}

@inproceedings{florence,
  author={Xiao, Bin and Wu, Haiping and Xu, Weijian and Dai, Xiyang and others},
  booktitle={2024 IEEE/CVF Conference on Computer Vision and Pattern Recognition (CVPR)}, 
  title={Florence-2: Advancing a Unified Representation for a Variety of Vision Tasks}, 
  year={2024},
  volume={},
  number={},
  pages={4818-4829},
  doi={10.1109/CVPR52733.2024.00461}}

\end{document}